\documentclass[letterpaper,twocolumn,10pt]{article}
\usepackage{usenix2019_v3}

\usepackage{tikz}
\usepackage{amsmath}
\usepackage{enumitem}

\usepackage{algpseudocode}
\usepackage[ruled,linesnumbered]{algorithm2e}
\usepackage{multirow}
\usepackage{appendix}
\usepackage{graphicx}
\usepackage{amssymb}
\usepackage{stfloats}
\usepackage{graphics}

\usepackage{filecontents}

\begin{document}

\date{}

\title{Model Stealing Attack against Graph Classification with Authenticity, Uncertainty and Diversity}
\author{Zhihao Zhu, Chenwang Wu, Rui Fan, Yi Yang, Zhen Wang, Defu Lian, Enhong Chen}

\maketitle

\begin{abstract}
Recent research demonstrates that GNNs are vulnerable to the model stealing attack, a nefarious endeavor geared towards duplicating the target model via query permissions. However, they mainly focus on node classification tasks, neglecting the potential threats entailed within the domain of graph classification tasks. Furthermore, their practicality is questionable due to unreasonable assumptions, specifically concerning the large data requirements and extensive model knowledge. To this end, we advocate following strict settings with limited real data and hard-label awareness to generate synthetic data, thereby facilitating the stealing of the target model. Specifically, following important data generation principles, we introduce three model stealing attacks to adapt to different actual scenarios: MSA-AU is inspired by active learning and emphasizes the uncertainty to enhance query value of generated samples; MSA-AD introduces diversity based on Mixup augmentation strategy to alleviate the query inefficiency issue caused by over-similar samples generated by MSA-AU; MSA-AUD combines the above two strategies to seamlessly integrate the authenticity, uncertainty, and diversity of the generated samples. Finally, extensive experiments consistently demonstrate the superiority of the proposed methods in terms of concealment, query efficiency, and stealing performance.

\end{abstract}

\section{Introduction}
Graph data has been extensively employed across various fields, including transportation networks~\cite{derrible2009network, zheng2020gman}, social networks~\cite{myers2014information, tang2010graph}, and chemical property prediction~\cite{cai2022fp, ma2020multi}.
To analyze the graph data, numerous graph-based machine learning (ML) models\cite{qiu2018network, tang2015line, tang2015pte} have been posited.
Graph neural networks (GNNs)~\cite{kipf2016semi, hu2020gpt, hamilton2017inductive}, representing one of the most cutting-edge graph-based paradigms, have gained noteworthy attention due to their superior performance in tasks such as node classification~\cite{zhou2019meta, wang2021bag}, link prediction~\cite{lei2019gcn, li2019mv} and graph classification~\cite{errica2019fair, lee2021learnable}.
Nonetheless, recent studies have underscored the susceptibility of GNNs to model stealing attacks in node classification tasks~\cite{shen2022model, wu2022model}. These attacks involve acquiring a replica of the model deployed within Machine Learning as a Service (MLaaS) systems, facilitated through query permissions. Such incursions harbor the potential to compromise the owner's intellectual property, escalate the vulnerability to adversarial activities~\cite{papernot2017practical, juuti2019prada}, and heighten susceptibility to membership inference attacks~\cite{shokri2017membership, zhu2023membership}, thereby undermining the credibility and privacy of the models.

\begin{table}[t]
\caption{The main differences between our work and related works~\cite{shen2022model, wu2022model}.}
\resizebox{\columnwidth}{!}{%
\begin{tabular}{l|lll}
\hline
                      & Related works         & Our framework          \\
\hline
Target   task         & node-level            & graph-level            \\
Attack capability     & unlimited data        & few real data          \\
     & intermediate output           & hard label             \\
\hline
\end{tabular}
}
\label{dif}
\end{table}

While prior work~\cite{shen2022model, wu2022model} has explored model stealing attacks targeting GNNs in node-level classification tasks, the threats of GNNs in graph-level classification tasks have remained unexplored. 
Moreover, previous research~\cite{shen2022model, wu2022model} assumed that attackers could access a substantial volume of real data, which comes from the same distribution with the target model's training data. They utilized query permissions on the target model to obtain prediction results for samples, which could be either confidence vectors of the classes likelihood or node representations. Subsequently, the attacker harnessed these data samples and prediction results to facilitate the replication of the target model by the clone model. However, the assumption of unrestricted access to source-similar data proves infeasible in graph-level tasks~\cite{wu2021adapting}. 
For example, in the case of AlphaFold~\cite{david2022alphafold}, its training data involves the resource-intensive constructions of biological networks, rendering it a formidable challenge for potential attackers to obtain a substantial amount of source-similar data. Additionally, not all models deployed on MLaaS systems possess the capability to furnish comprehensive predictive results. Most MLaaS systems~\cite{sanyal2022towards, wang2022black} merely offer categorical labels for samples (e.g., the image depicts a cat or a dog, rather than the probabilities of the image belonging to various categories). These challenges force us to urgently study more practical model stealing attacks for graph classification.

To this end, we assume that attackers can only access a limited amount of source-similar data to the target model and resolve the issue of insufficient real samples through the generation of new samples. We propose two principles for the generation process of new samples to ensure the attack's imperceptibility and effectiveness. 
\begin{itemize}[leftmargin=*]
    \item \textbf{Principle 1.} \textit{\textbf{Authenticity}. Generated samples should exhibit minimal differences from real samples while maintaining an adequate level of authenticity.}
    \item \textbf{Principle 2.} \textit{\textbf{Query Value}. Generated samples should possess sufficient query value to enhance the effectiveness of the target model simulation, that is, more informative for mimicking the target model's behavior.}
\end{itemize}
It is evident that the constraints of Principle 1 preserve the requisite covert nature of generated samples, thereby eluding the detection mechanisms of the target model. Simultaneously, adherence to the ideal properties of Principle 2 augments the efficacy of the attack, allowing for heightened performance within the confines of a restricted query budget. Grounded in these two fundamental principles, we develop three model stealing attacks, MSA-AU, MSA-AD, and MSA-AUD.

\textbf{MSA-AU emphasizes authenticity and uncertainty.} 
Drawing inspiration from the doctrines of active learning~\cite{settles2009active, ren2021survey}, it deems samples yield augmented uncertainty in the predictions of the extant model as possessing elevated query worth. This elevation in query value stems from the fact that heightened uncertainty suggests the proximity of these samples to the model's decision threshold. As a consequence, affixing labels to these samples characterized by substantial uncertainty can substantially assist the model in rectifying prior fallacious predictions. Building on it, we adopt the idea of adversarial attacks to introduce perturbations in the accessible real samples, aiming to get samples with higher uncertainty. Notably, we focused on the perturbations on the graph's adjacency matrix rather than node features, which avoids generating unreasonable node features that undermine the requirements of Principle 1. To further maintain authenticity, attackers can modify only a minority of adjacency matrix positions, guided by importance scores from the clone model.  By limiting the perturbation, the generated synthetic samples strike a balance between authenticity and uncertainty. 

\textbf{MSA-AD prioritizes authenticity and diversity.} 
The uncertain samples generated by MSA-AU rely on the clone model, which may lead to diminishing differences among the generated samples as the clone model converges. Consequently, this leads to assailants persistently soliciting redundant samples, depleting the query budget ineffectually. Drawing inspiration from the concept of active learning~\cite{brinker2003incorporating, yang2015multi}, we introduce diversity as a measure of the query worth of a batch of samples. Accordingly, we introduce an alternative model-independent sample generation strategy: MSA-AD, to prevent excessive similarity among the generated samples. In MSA-AD, we employ Mixup~\cite{wang2021mixup, yoo2022model} to derive new samples by blending multiple samples. Specifically, two randomly selected genuine samples serve as the foundation, from which MSA-AD selects a small subset of nodes from each to form corresponding induced subgraphs. Subsequently, MSA-AD exchanges the topological structures and node attributes of these induced subgraphs. Given the minimal number of modified nodes, the statistical disparities between the generated graphs and their original counterparts remain subtle. Moreover, MSA-AD applies feature copying to modify node attributes, thereby mitigating the risk of inconsistencies in feature content. To further amplify the diversity, we strive to pair the original graph with diverse graphs as extensively as possible in various iterations.

\textbf{MSA-AUD concurrently considers authenticity, uncertainty, and diversity within the generated sample ensemble.} Specifically, we first apply the MSA-AU strategy to the original samples and subsequently use MSA-AD on the results. Lastly, we feed the twice-modified new samples into the target model for querying to aid in the training of the clone model. Given that both MSA-AU and MSA-AD address the issue of attack authenticity, the disparities between the new samples and the original samples remain minimal. Benefiting from the advantages of the above two methods, MSA-AU ensures the uncertainty of the new samples, while the model-independent MSA-AD guarantees diversity within the sample collection.

Our contributions can be summarized as follows:
\begin{itemize}[leftmargin=*]
    \item We are the first to explore model stealing attacks for graph classification tasks. Our work involves adopting stringent assumptions regarding the attacker's capabilities, including limited real data and exclusive availability of hard labels. The proposed methods based on these assumptions enjoy both effectiveness and practicality.
    \item We introduce important principles governing the sample generation. In alignment with these principles, we proffer three strategic methodologies aimed at ensuring the authenticity, uncertainty, and diversity of the generated samples.
    \item Extensive experiments demonstrate that our attack method is covert, high-performing, and efficient in various scenarios. Even the latest defense methods cannot fully withstand our attack. In addition, through experimental analysis, we suggest several viable directions for defending against model extraction attacks in graph classification tasks.
\end{itemize}

\section{Target Model}

\begin{table}[t]
\caption{Summary of the notations.}
\begin{tabular}{l|l}
\hline
Notation                 & Description               \\ \hline
$g=(\mathcal{V},\textbf{A},\textbf{X})$              & Graph                     \\
$g^s$                             & A subgraph of $g$         \\
$\textbf{A}$                      & Adjacement matrix         \\
$\textbf{X}$                      & Node features             \\
$y_g$                & Label of graph $g$    \\
$\mathcal{M}_T$          & The target model                             \\ 
$\mathcal{M}_C$          & The clone model                              \\
$\mathcal{P}$            & The sample pool                              \\ 
$\mathcal{D}_{0}$        & The Initial dataset                          \\
$N$                    & The number of iterations               \\
$\alpha$        & Modification rate of adjacency matrix\\
$\gamma$                & Proportion of mixed nodes\\
\hline
\end{tabular}
\label{notation}
\end{table}

In this section, we first introduce the target model.
Then we characterize the attacker’s goal, background knowledge, and attack setting.
Table~\ref{notation} provides a convenient summary of the notations introduced in this research.
\subsection{Graph Neural Network (GNN)}
In recent years, GNNs have been frequently employed for node classification, link prediction, and graph classification due to their exceptional performance.
The effectiveness of GNNs is attributable to the message propagation and aggregation mechanism, which may effectively combine information from neighboring nodes.
Specifically, a single graph convolutional layer of a GNN can transmit the knowledge of its first-order neighbors to a node, and there are numerous such layers in GNN.
After $k$ iterations of message passing, a node can capture abundant information from its $k$-hop neighbors.
Formally, the message passing and aggregating of a graph convolutional layer in a GNN can be defined as the following:
\begin{equation}
\textbf{h}_v^l = AGGREGATE(\textbf{h}_v^{l-1}, MSG(\textbf{h}_v^{l-1}, \textbf{h}_u^{l-1})), u \in \mathcal{N}_v,
\end{equation}
where $\textbf{h}_v^{l-1}$ and $\textbf{h}_u^{l-1}$ indicate the $l-1$ layer embedding of node $v$ and $u$.
$AGGREGATE$ and $MSG$ are the corresponding aggregate function and message-passing function.
$\mathcal{N}_v$ contains $v$'s first-order neighbors.

Once trained, each node of the graph can be represented as an embedding vector, which integrates both graph structure and node features.
In graph classification, the model's goal is to classify the whole graph.
Therefore, we need to extract graph embedding from the node embeddings, which is called the pooling operation at the graph level.
\begin{equation}
\textbf{h}_g^l = Pooling(\textbf{h}_v^l), v \in g.
\end{equation}
For a $k$-layer GNN, the objective function in graph classification can be defined as:
\begin{equation}
\max \limits_{\theta} \sum_{g\in \mathcal{G}_{T}}^{}I(f_\theta(\textbf{h}_g^0, \textbf{h}_g^1, ..., \textbf{h}_g^k)=y_g),
\end{equation}
where $\textbf{h}_g^l$ and $y_g$ are $l$-th layer's embedding vector and label of graph $g$.
$\mathcal{G}_{T}$ is the graph training set.

\subsection{Attacker’s Goal}
Model stealing attacks involve creating a clone model $\mathcal{M}_C$ to mimic the behavior of the target model $\mathcal{M}_T$.
There are two metrics to measure the performance of model stealing attacks, one is the accuracy of $\mathcal{M}_C$, and the other is fidelity.
Accuracy is used to estimate whether the prediction of $\mathcal{M}_C$ matches the ground truth label, while fidelity measures the degree of agreement between the predictions of $\mathcal{M}_C$ and $\mathcal{M}_T$.
\begin{equation}
Accuracy = \frac{1}{|\mathcal{G}_{T}|} \sum_{g\in \mathcal{G}_{T}}^{}I(\mathcal{M}_C(g)=y_g).
\end{equation}
\begin{equation}
Fidelity = \frac{1}{|\mathcal{G}_{T}|} \sum_{g\in \mathcal{G}_{T}}^{}I(\mathcal{M}_{C}(g)=\mathcal{M}_{T}(g)).
\end{equation}

Evaluating model stealing attacks using fidelity is a common practice in recent works~\cite{wang2022dualcf, dubey2022high}.
In this setting, the attackers do not force the clone model to perform well, but expect it to mimic the target model.
It means that even if the target model makes wrong predictions, the clone model should imitate them.
A well-imitated clone model is profitable for the attackers to implement subsequent attacks, such as adversarial attacks~\cite{papernot2017practical, juuti2019prada} and membership inference attacks~\cite{shokri2017membership, zhu2023membership}.
Therefore, we adopt \textbf{fidelity} to evaluate our method.

\subsection{Attacker’s Capability}
Previous works assumed that the attacker can obtain enough data which comes from the same distribution of the target model's training data, namely target data. However, this assumption is often unrealistic in practical applications. In this paper, we assume that the attacker only has access to a small amount (10\% or even 1\% in our experiments) of the target data. In addition, we assume that the attacker has query access to the target model, but the attacker cannot obtain intermediate results (node embeddings or confidence vectors) of the target model as in previous works~\cite{shen2022model}. For a given query sample, the target model will simply return the target model's prediction of the class for that sample, namely the hard label. Finally, we assume that the attacker knows the architecture of the target model, an assumption that will be relaxed in section \ref{unware_exp}.

\subsection{Attack Setting}
In this article, we use synthetic samples to mitigate the problem of insufficient target data available. We synthesize new data by making a small number of modifications to the original data. In order to make the synthetic samples more realistic and our attacks less perceptible, we make strict constraints on the range and scale that the attacker can modify. Specifically, we discuss model stealing attacks against graph classification tasks under two scenarios. One scenario is when the attacker can only slightly modify the topology of the original graph, the adjacency matrix. In the other scenario, the attacker can not only modify the adjacency matrix but also replace the node features with other features of the same distribution as it. Note that we do not modify the node features here, because modifying node features is prone to contradictory situations. 

\section{Model Stealing Attack}

\subsection{Overview}
\begin{figure}[t]
  \centering
  \includegraphics[width=\linewidth]{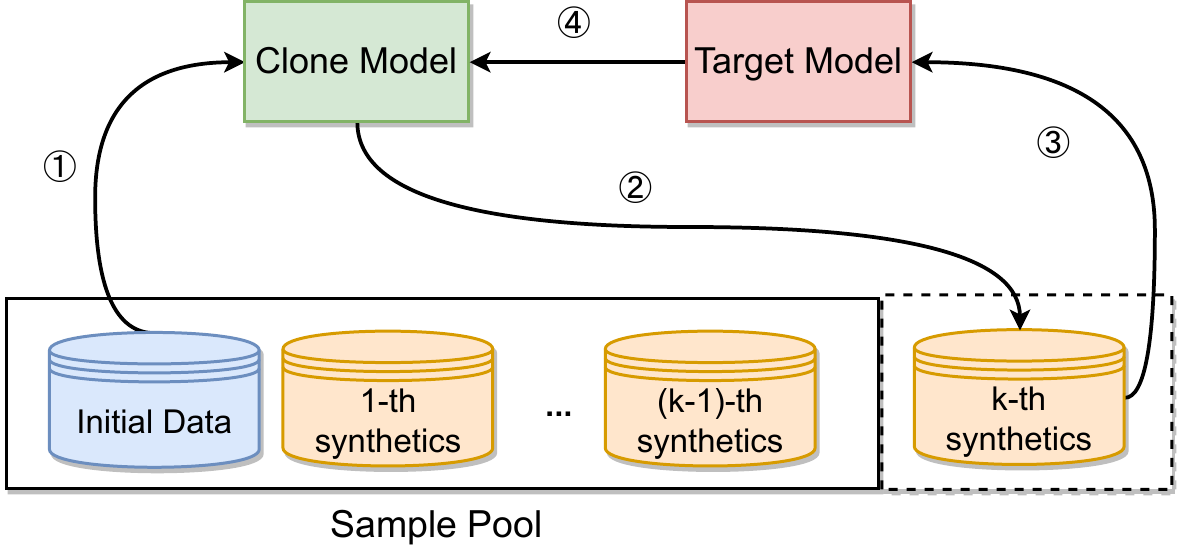}
  \caption{The workflow of our framework. Each round begins with the generation of new synthetic data by applying model-related or model-agnostic methods. We then query new samples against the target model, whose predictions will be utilized to boost the clone model's imitation. }
  \label{workflow}
\end{figure}
Our framework consists of three components: (i) the target model $\mathcal{M}_T$, (ii) the clone model $\mathcal{M}_C$, and (iii) the sample pool $\mathcal{P}$.
The target model provides a query interface for the public and returns hard labels for query graphs. 
The goal of the clone model is to imitate $\mathcal{M_T}$'s predictions, including both correct and incorrect decisions. The expansion of $\mathcal{P}$ is the core of the framework, where the attacker synthesizes authentic and valuable data to persuade $\mathcal{M_C}$ to imitate $\mathcal{M_T}$. 

The workflow of our framework is shown in Fig~\ref{workflow}.
We initialize the sample pool with a small portion of the real data $\mathcal{D}_{0}$(initial data) and pretrain the clone model $\mathcal{M}_{C}$ on $\mathcal{D}_{0}$. 
Each cycle begins with synthesizing new samples through the model-related method (MSA-AU), model-agnostic method (MSA-AD), or both(MSA-AUD).
Next, we query the target model for these synthetic examples and add them to the pool of samples ($\mathcal{P}$).
Finally, we utilize the samples in $\mathcal{P}$ and the corresponding hard labels predicted by the target model to train $\mathcal{M}_{C}$.
Alternating iterations of sample creation and training of $\mathcal{M}_{C}$ eventually condition $\mathcal{M}_{C}$ to be an excellent duplicate of $\mathcal{M}_{T}$.
In the following subsections, we introduce three strategies MSA-AU, MSA-AD, and MSA-AUD, which guarantee authenticity and uncertainty, authenticity and diversity, all three characteristics of generated samples, respectively.

\subsection{Model Stealing Attacks with Authenticity and Uncertainty}
\subsubsection{Optimization objective based on uncertainty}
Inspired by active learning~\cite{settles2009active, ren2021survey}, we can employ uncertainty as a metric to assess the value of generated samples.
The basic idea of active learning is to promote a machine learning system to attain greater accuracy with fewer labeled training instances.
In the context of active learning, uncertainty plays a crucial role in evaluating the knowledge content within samples. 
Samples with higher uncertainty are more challenging for models to differentiate and, as a result, are deemed more valuable. 
Commonly used metrics for quantifying uncertainty include: margin of confidence (Margin), minimized maximum confidence (Max), and entropy of confidence vector (Entropy), which have the following formula:
\begin{equation}
    Margin(g, \mathcal{M}) = P_{\mathcal{M}}(y_2|g) - P_{\mathcal{M}}(y_1|g)
\end{equation}
\begin{equation}
    Max(g, \mathcal{M}) = -P_{\mathcal{M}}(y_1|g)
\end{equation}
\begin{equation}
    Entropy(g, \mathcal{M}) = -\sum_{y \in C_g} P_{\mathcal{M}}(y|g)  * log(P_{\mathcal{M}}(y|g)),
\end{equation}
where $C_g$ represents all possible classes to which graph $g$ can belong, and $P_{\mathcal{M}}(y|g)$ is the likelihood that the model $M$ believes graph $g$ belongs to class $y$. $y_1$ and $y_2$ are the two classes in which the model considers graph $g$ to belong to the largest and the second largest probability. The greater the value of these metrics, the greater the sample's uncertainty.

Diverging from conventional active learning, which entails \textbf{selecting} samples with higher uncertainty from real samples, we opt to modify genuine samples based on their predictions by the clone model, thereby \textbf{generating} new samples with heightened uncertainty.
Querying the generated samples with high uncertainty and directing the clone model with the relevant responses of the target model can assist the clone model in better correcting the discrepancies between the target model and the clone model.
Typically, genuine samples undergo a forward pass as inputs into the clone model. To create new samples with increased uncertainty, we treat the samples as parameters, while keeping model parameters fixed and using uncertainty as the optimization objective. Employing both forward and backward passes, gradients of the loss (uncertainty) with respect to parameters (the genuine samples) can be computed. Modifying samples in the direction opposite to the gradient yields new samples with greater uncertainty in the clone model.

\begin{algorithm}[h]
    \label{AU_gen}
    \caption{$AU\_gen$}
    \KwIn{An original sample $g$; The clone model $\mathcal{M}_{C}$; Modification rate of adjacency matrix $\alpha$;}
    \KwOut{A new sample $g'$;}
    $\mathcal{V}, A, X\xleftarrow{}g; \hfill //$The adjacency matrix and features\\
    $U_g\xleftarrow{}Uncertainty(g, \mathcal{M}_{C})$\;
    $Grad_A\xleftarrow{}\frac{\partial U_g}{\partial A}$\;
    \For{$i \in$ A's rows, $j \in$ A's columns}
    {
        \If{($A[i,j]$ is 0 and $Grad_A[i,j] < 0$) or\\ ($A[i,j]$ is 1 and $Grad_A[i,j] > 0$)}{
            $Grad_A$[i,j] $\xleftarrow{} |Grad_A[i,j]|$\;
        }
        \Else {
            $Grad_A$[i,j] $\xleftarrow{}$ 0\;
        }
    }
    {$M\xleftarrow{}Argmax\_desc(Grad_A)$}\;
    $A'\xleftarrow{}A$\;
    \For{$(i,j) \in M[0:\alpha \dot |M|]$}
    {
        $A'[i,j]\xleftarrow{}1-A'[i,j]$\;
        $g'\xleftarrow{}(\mathcal{V}, A',X)$\;
        $U_{g'}\xleftarrow{}Uncertainty((A',X), \mathcal{M}_{T})$\;
        \If{$U_{g'} > 0.1$}{break}
        \# Authenticity constraint
        
        \If{$|Statistics(g)-Statistics(g')|>0.05$}{break} 
    }
\end{algorithm}

However, this approach encounters two notable challenges. 
On one hand, modifying features through this method can lead to results devoid of meaning. For instance, when a node represents a user in a social network, it is conceivable to create a user who is seven years old and has been working for more than 20 years, which is obviously unreliable. The target model might flag these samples as malicious queries, adversely affecting stealing performance. 
On the other hand, this modification technique is only applicable to continuous variables and is unsuitable for altering discrete adjacency matrices. Even when adjacency matrices are relaxed to continuous variables, followed by the addition of perturbation and reconversion to discrete variables, this method may still yield graphs that differ significantly from the originals, thus going against our first principle.

To address these two challenges, we constrain the attacker to modify only a limited number of positions within the adjacency matrix, aiming to generate samples with higher authenticity and uncertainty. We evaluate the potential of each position in the adjacency matrix for enhancing sample uncertainty by the sample's gradient under the current clone model. In this study, we assume that the adjacency matrix lacks multiple edges, meaning the adjacency matrix values are confined to $\{0,1\}$. Consequently, there is no need to introduce negative perturbation for positions with a value of 0 or positive perturbation for positions with a value of 1. Therefore, we denote the potential score of these positions as 0 and use the absolute value of the gradient as the potential score for the other positions.
Portions with potential score exceeding zero signify that modifying these positions can increase sample uncertainty. We greedily select $\alpha$ positions with higher importance and flip the values of the corresponding positions in descending order of importance. The impact of the $\alpha$ on the attack model is discussed in Section \ref{Modified_Edge_Numbers}. Given the minimal alterations to edges, the generated graph closely resembles the original, thereby ensuring the imperceptible of the attack method.
To ensure that the generated new samples do not exhibit excessive differences in uncertainty compared to the original graph, we assess uncertainty of the new graph after every modification. If its uncertainty surpasses a predefined threshold, we terminate the algorithm.

\subsubsection{Authenticity constraint}
While constraining the number of modifications to the adjacency matrix can to some extent ensure the authenticity of the generated graph, we are unable to measure whether certain statistical features of the graph, such as degree distribution and the number of triangles, have undergone significant changes. Therefore, when greedily selecting modification positions, we introduce additional authenticity constraints to our attack method. After each modification, we assess the statistical difference between the new graph and the original graph. If this difference exceeds a predefined threshold, the attack process terminates prematurely. For example, in the case of degree distribution, the authenticity constraint stipulates that the difference in degree distribution between the generated and original graphs should be less than 0.05. In section \ref{real}, we observe that these authenticity constraints are effective in maintaining the similarity between the generated graph and the original graph across multiple statistical features. 

The algorithm that combines uncertainty and authenticity, namely MSA-AU, is a model-related sample generation method. Algorithm \ref{AU_gen} provides a detailed description of the sample generation steps in the MSA-AU algorithm, explicitly outlining the process that ensures both uncertainty and authenticity. $Argmax\_desc$ provides a collection of indices. The indices in the set are sorted descently according to the potential score of the input matrix indicated by the indices.

\subsection{Model-independent Sample Generation Strategy}
\begin{figure}[htbp]
    \centering
    \includegraphics[width=0.8\linewidth]{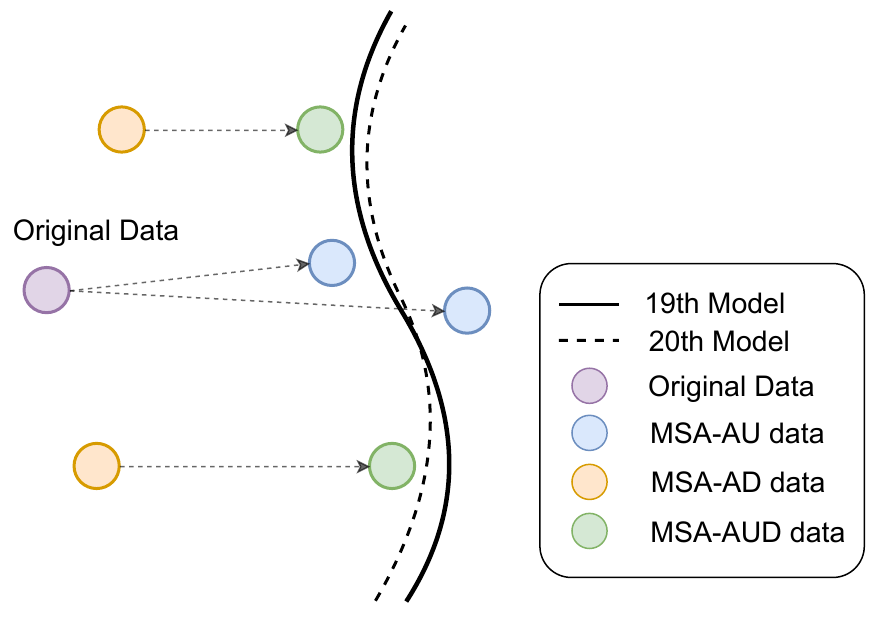}
    \caption{Difference between MSA-AU, MSA-AD and MSA-AUD.}
    \label{3algo}
\end{figure}
In MSA-AU, the generated, label-worthy samples are related to the current clone model. While this model-related generation method is effective in obtaining samples with increased uncertainty, the differences between these samples decrease gradually with the convergence of model training across multiple iterations. This trend is intuitively depicted in Fig \ref{3algo}, where the solid line represents the decision boundary of the clone model in the 19th round, and the dashed line represents the decision boundary in the 20th round. It can be observed that in the later stages of model training, the change in the clone model's decision boundary is minimal, resulting in limited differences between the MSA-AU generated samples in the 19th and 20th rounds. Although both sets of samples exhibit high uncertainty, labeling them simultaneously does not provide satisfactory assistance to the clone model. 
In addition to ensuring the uncertainty and authenticity of new samples, we also aim to introduce diversity among these samples. This concept is akin to the diversity selection criteria in active learning~\cite{brinker2003incorporating, yang2015multi}. In active learning, besides selecting samples solely based on uncertainty, diversity is typically used to assess the sample set to avoid redundancy. Similarly, to achieve superior attack performance with a minimal number of queries, we must ensure diversity among the generated sample set while synthesizing samples.

Based on this, we propose another model-independent sample generation strategy, which involves generating new samples through the fusion of real samples. Since the fusion process is model-independent, it does not lead to a loss of diversity in the sample pool as the clone model converges. Specifically, we begin by randomly selecting original graphs $g_1$ and $g_2$ from the initial sample pool. Subsequently, we determine the proportion of nodes to be modified and obtain corresponding induced subgraphs $g^s_1$ and $g^s_2$ for $g_1$ and $g_2$ based on this proportion. Finally, we establish a one-to-one mapping relationship between the nodes in $g^s_1$ and $g^s_2$. Through this one-to-one mapping, we exchange the topological structure and node features of $g^s_1$ and $g^s_2$, resulting in modified graphs denoted as $g_1'$ and $g_2'$. This process is referred to as mixup~\cite{wang2021mixup, yoo2022model} for graph data, an emerging graph data augmentation method. In our paper, the attack model using this sample generation strategy is named as MSA-AD.

\begin{algorithm}[h]
    \label{AD_gen}
    \caption{$AD\_gen$}
    \KwIn{An original sample $g_1$; the Initial Dataset $\mathcal{D}_0$; Proportion of mixed nodes $\gamma$}
    \KwOut{A new sample $g_1'$;}
    $g_2\xleftarrow{}randomly\_sample(\mathcal{D}_0/ \{g_1\})$\;
    $v_1\xleftarrow{}randomly\_sample(g_1);\hfill //$Select a Vertex\\
    $v_2\xleftarrow{}randomly\_sample(g_2)$\;

    $g^s_1\xleftarrow{}Expand\_from(v_1, \gamma);\hfill //$Induce Subgraph\\
    $g^s_2\xleftarrow{}Expand\_from(v_2, \gamma)$\;
    $PageRank(g^s_1)\leftrightarrow PageRank(g^s_2)$\;
    $\hfill //$One-to-one Coorespondence\\
    $g_1'\xleftarrow{} Replace(g_1, g^s_1, g^s_2)$\;
\end{algorithm}

On one hand, due to the limited proportion of adjacent matrix (just induced subgraph) being modified, the differences in statistical features between the generated and original graphs are minimal. On the other hand, the copied node features in MSA-AD are drawn from nodes in other graphs that belong to the same distribution, ensuring that no feature content contradictions arise. This method of copying features is also considered to have a higher level of imperceptibility in several studies~\cite{fan2021attacking, wu2021triple}.

The core of MSA-AD lies in establishing a one-to-one mapping of nodes in subgraphs. To ensure minimal differences in statistical features between the generated and original graphs, we aim to create a one-to-one correspondence for nodes that play significant roles in the graph. For instance, we do not want nodes with higher degrees in graph $g_1$ to be mapped to nodes with lower degrees in graph $g_2$. Therefore, we use the PageRank~\cite{brin1998pagerank} algorithm to calculate the importance of each node in graphs $g_1$ and $g_2$ and link the mapping relationship of nodes in the subgraphs to their importance. Nodes with higher importance in $g_1$ are mapped to nodes with higher importance in $g_2$. The importance of each node in the original graph can be pre-computed, eliminating the need for repetitive importance calculations in different rounds of expanding the sample pool. Furthermore, to ensure a greater diversity of generated samples, we require different pairings for each original graph in various rounds. In other words, each pairing of graph $g_1$ with graph $g_2$ will appear at most once. 

\begin{table}[]
\centering
\caption{Comparison of MSA-AU and MSA-AD}
\begin{tabular}{c|cc}
\hline
\textbf{}                     & {MSA-AU} & {MSA-AD} \\ \hline
{Model-related}        & $\checkmark$    &                 \\ \hline
{Modify Adjacent Matrix}    & $\checkmark$    & $\checkmark$    \\ \hline
{Modify Node features} &                 & $\checkmark$    \\ \hline
\end{tabular}
\label{AU-AD-comp}
\end{table}

Table \ref{AU-AD-comp} provides a detailed comparison between MSA-AU and MSA-AD. MSA-AU is model-related and considers sample uncertainty and authenticity, with constraints limited to modifying adjacency matrices. MSA-AD, on the other hand, is model-independent, considers sample set diversity and authenticity, and can simultaneously modify adjacency matrices and copy node features.
Compared with MSA-AU, the difference between MSA-AD generated data in the 19th and 20th round is significant, as clearly depicted in Fig \ref{3algo}. As MSA-AD is model-agnostic, the MSA-AD data does not have any connection to the 19th and 20th models.

For a detailed procedure of MSA-AD's sample-generating process, please refer to Algorithm \ref{AD_gen}. 
In practice, subgraphs are generated by randomly selecting a central vertex and expanding from it to form a connected cluster comprising $\gamma$ rate of the total vertices. The discussion of $\gamma$'s value can be found in Section \ref{mix}. Afterward, PageRank is used to establish a one-to-one correspondence, and the final new sample is created by switching subgraphs accordingly.

\begin{algorithm}[t]
    \label{algo}
    \caption{MSA-AUD}
    \KwIn{The trained target model $\mathcal{M}_{T}$; The initial dataset $\mathcal{D}_0$; The number of iterations N}
    \KwOut{The trained clone model $\mathcal{M}_{C}$;}
    Select and Initialize $\mathcal{M}_{C}$(e.g. GCN) \;
    $\mathcal{L}_0\xleftarrow{}Query(\mathcal{M}_{T}, \mathcal{D}_0)$ \;
    Initialize the sample pool $\mathcal{P}$ with $(\mathcal{D}_0, \mathcal{L}_0)$ \;
    Train $\mathcal{M}_{C}$ with $\mathcal{P}$\;
    \For{$i \in {1,...,N}$}
    {
        $\mathcal{D}_i\xleftarrow{}\{\}$\;
        \For {$g \in \mathcal{D}_0$}
        {
            $g'\xleftarrow{}AD\_gen(g, \mathcal{D}_0)$\;
            $g''\xleftarrow{}AU\_gen(g', \mathcal{M}_{C})$\;
            $\mathcal{D}_i\xleftarrow{}\mathcal{D}_i\bigcup\{g''\}$\;
        }
        $\mathcal{L}_i\xleftarrow{}Query(\mathcal{M}_{T}, \mathcal{D}_i)$ \;
        $\mathcal{P} = \mathcal{P} \cup {(\mathcal{D}_i, \mathcal{L}_i)}$ \;
        Train $\mathcal{M}_{C}$ with $\mathcal{P}$\;
    }
\end{algorithm}

\subsection{MSA-AUD}
In this section, we introduce MSA-AUD, a strategy that combines the advantages of MSA-AU and MSA-AD, simultaneously incorporating authenticity, uncertainty, and diversity during the process of expanding the sample pool.

Specifically, in the process of expanding the sample pool, we initially employ the MSA-AD method to blend the original graphs $g_1$ and $g_2$, resulting in modified graphs $g_1'$ and $g_2'$. Subsequently, building upon this, we execute a model-dependent generation process on $g_1'$, akin to the steps in MSA-AU, to obtain generated sample $g_1''$. As both modification processes constrain the differences between the generated graphs and the original ones, the authenticity of the generated samples is assured. Fig \ref{3algo} illustrates the distinctions between MSA-AU, MSA-AD, and MSA-AUD. It can be observed that while MSA-AU can generate synthetic samples with higher uncertainty based on the original data, it is susceptible to losing diversity due to the influence of the clone model. On the other hand, MSA-AD, although model-independent and avoiding the generation of redundant samples, cannot guarantee individual sample uncertainty, i.e., label worthiness. MSA-AUD combines the advantages of the formers, ensuring the authenticity, uncertainty, and diversity of the synthetic sample set. 

The complete workflow of MSA-AUD is detailed in Algorithm \ref{algo}.
In the algorithm, the Query function receives a target model and a dataset. It inputs the dataset into the target model and returns the predictions. $AD\_gen$ and $AU\_gen$ refer to Algorithm \ref{AD_gen} and Algorithm \ref{AU_gen}, respectively.

\textbf{Diversity visualization}.
\begin{figure}[t]
  \centering
  \includegraphics[width=\linewidth]{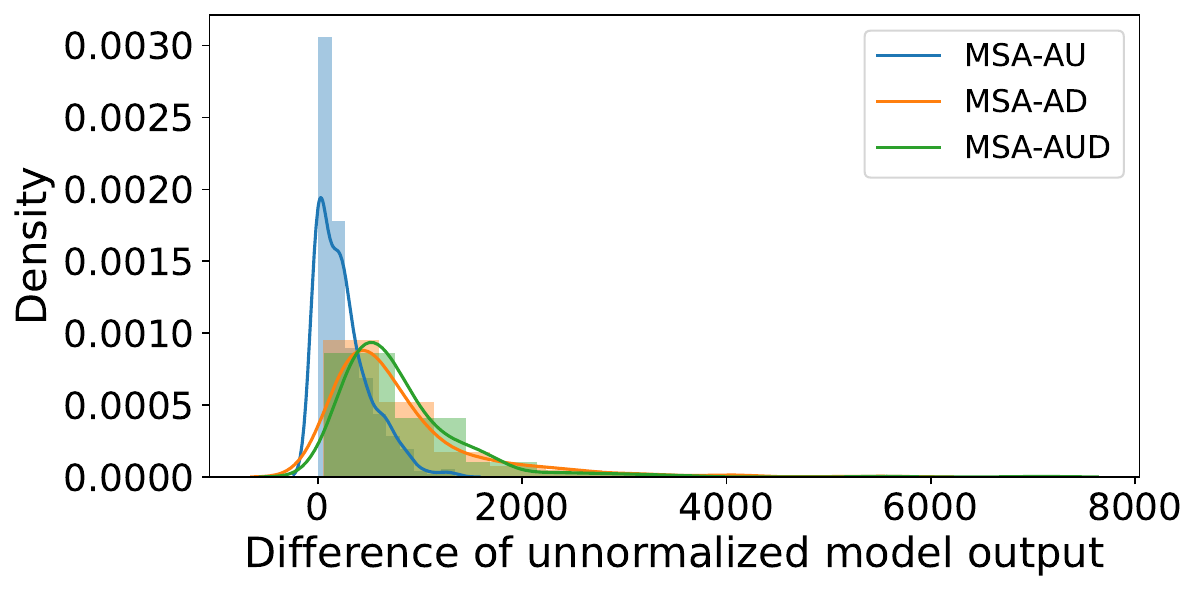}
  \caption{Diversity visualization.}
  \label{visual}
\end{figure}
In order to investigate whether the introduction of mixup genuinely enhances data diversity, a statistical analysis is conducted on the target model. We synthesize new samples based on the initial data for three different methods at the 19th and 20th iterations. These synthetic samples are subsequently input into the target model. The squared sum of the model output vectors' differences between two consecutive iterations is computed as a measure of the dissimilarity between the two classes of data. The results are visualized using histograms and fitted with probability density functions, as presented in Fig.\ref{visual}.

From the observations in the figure, it can be noted that the dissimilarity values for MSA-AU are relatively concentrated and centered around zero, indicating that the samples generated by MSA-AU in two consecutive iterations exhibit minimal differences. In contrast, the MSA-AD algorithm, primarily focused on ensuring diversity, shows a peak around 500, signifying a substantial increase in dissimilarity compared to MSA-AU. Meanwhile, the MSA-AUD algorithm, combining the strengths of the first two algorithms, exhibits similar diversity results to MSA-AD. These results serve as compelling evidence to demonstrate the effectiveness of our approach in enhancing sample diversity.

\section{Evaluation}
In this section, we evaluate several attack methods against three Graph Neural Network (GNN) models on multiple datasets. We begin by introducing our experimental setup. Following the experimental setup, we present the attack performance and imperceptibility of different algorithms in Sections \ref{mainre} and \ref{imper}, respectively. Then we assess the performance of attack methods with varying knowledge and parameters in Sections \ref{knowledge} and \ref{param}, respectively. Furthermore, we discuss the performance of our attack methods under defense strategies in Section \ref{defense}. Finally, we discuss the feasible defense directions based on the experimental results in Section \ref{discuss}.

\subsection{Setup}
\subsubsection{Dataset}
\begin{table}[t]
\caption{The statistics of the datasets}
\scalebox{0.9}{
\begin{tabular}{l|rrrr}
\hline\noalign{\smallskip}
Dataset       & \#Graphs & \#Classes & \#Nodes & \#Edges \\
\noalign{\smallskip}\hline\noalign{\smallskip}
ENZYMES        & 600      & 6         & 19580     & 74564     \\
COIL-DEL       & 3900     & 100       & 83995     & 423048    \\
NCI1           & 4110     & 2         & 122747    & 265506    \\
TRIANGLES      & 45000    & 10        & 938438    & 2947024   \\
\noalign{\smallskip}\hline
\end{tabular}}
\label{dataset}
\end{table}

We evaluate our attack algorithms on four datasets: ENZYMES, COIL-DEL, NCI1, and TRIANGLES. The statistical characteristics of these datasets are summarized in Table~\ref{dataset}. For each dataset, we randomly select 80\% of the graphs for training the victim model. We assume that the attacker can only access 10\% of the real data to Initialize the sample pool for cloning. Specifically, for the TRIANGLES dataset, we assume that the attacker can only use 1\% of the real data.
\begin{itemize}[leftmargin=*]
\item \textbf{ENZYMES}~\cite{borgwardt2005protein, schomburg2004brenda} included 600 proteins from each of the 6 Enzyme Commission top-level enzyme classes (EC classes) and the goal was to correctly predict enzyme class membership for these proteins.
\item \textbf{COIL-DEL}~\cite{riesen2008iam, nene1996columbia} contains 3900 graphs which are converted from images by the Harris corner detection algorithm~\cite{harris1988combined} and Delaunay triangulation~\cite{lee1980two}. Each node of these graphs is given a feature vector that comprises the node's position.
\item \textbf{NCI}~\cite{shervashidze2011weisfeiler} is a biological dataset used for anticancer activity classification. In this dataset, each graph represents a compound, with nodes and edges representing atoms and chemical bonds, respectively. The graph labels indicate whether the corresponding compounds exhibit positive or negative effects on lung cancer cells.
\item \textbf{TRIANGLES}~\cite{knyazev2019understanding}. Counting the number of triangles in a graph is a common task that can be solved analytically but is challenging for GNNs. This dataset contains ten different graph classes, corresponding to the number of triangles in each graph in the dataset. 
\end{itemize}

\subsubsection{Target Model} We choose three classic graph neural networks SAGE, GCN, and GIN to verify the performance of different algorithms.
\begin{itemize}[leftmargin=*]
\item \textbf{SAGE}~\cite{hamilton2017inductive} collects and combines messages from node's neighbors.
Due to the fact that it only requires the local topology of nodes, it can be easily implemented in large-scale graphs.
\item \textbf{GCN}~\cite{kipf2016semi} is one of the most widely used GNNs, simplifies the GNN model using ChebNet~\cite{defferrard2016convolutional}'s first-order approximation. 
Compared to conventional node embedding techniques~\cite{qiu2018network, tang2015line}, it performs much better in graph representation learning tasks.
\item \textbf{GIN}~\cite{xu2018powerful} analyzes the upper bound on GNNs' performance and proposes an architecture to satisfy this bound.
It is widely employed in graph classification tasks and is gradually becoming an essential baseline.
\end{itemize}

\begin{table*}[]
\centering
\caption{Attack performance(Fidelity) on various datasets. The best results are highlighted in bold.} 
\begin{tabular}{l|ccc|ccc|ccc|ccc}
\hline\noalign{\smallskip}
Dataset & \multicolumn{3}{c|}{ENZYMES} & \multicolumn{3}{c|}{COIL-DEL} & \multicolumn{3}{c|}{NCI1} & \multicolumn{3}{c}{TRIANGLES} \\
\noalign{\smallskip}\cline{1-13} \noalign{\smallskip}
Model   & SAGE    & GCN     & GIN     & SAGE     & GCN     & GIN     & SAGE   & GCN    & GIN    & SAGE     & GCN      & GIN     \\
\noalign{\smallskip}\hline\noalign{\smallskip}
MSA-Real    & 0.435   & 0.760   & 0.202   & 0.228    & 0.211   & 0.222   & 0.843  & 0.911  & 0.818  & 0.347    & 0.425    & 0.521   \\
JbDA    & 0.494   & 0.840   & 0.452   & 0.477    & 0.515   & 0.458   & 0.912  & 0.944  & 0.861  & 0.559    & 0.685    & 0.633   \\
T-RND    & 0.546   & 0.827   & 0.448   & 0.465    & 0.489   & 0.461   & 0.910  & 0.944  & 0.861  & 0.565    & 0.696    & 0.625   \\
\textbf{MSA-AU}  & \textbf{0.577}   & \textbf{0.854}   & \textbf{0.452}   & \textbf{0.590}    & \textbf{0.616}   & \textbf{0.589}   & \textbf{0.944}  & \textbf{0.966}  & \textbf{0.914}  & \textbf{0.609}    & \textbf{0.700}    & \textbf{0.739}   \\
\noalign{\smallskip}\hline\noalign{\smallskip}
Random  & 0.519   & 0.850   & 0.346   & 0.497    & 0.471   & 0.516   & 0.873  & 0.935  & 0.818  & 0.475    & 0.647    & 0.642   \\
MSA-AD   & 0.669   & 0.890   & 0.502   & 0.650    & 0.691   & 0.665   & 0.947  & 0.963  & 0.909  & 0.623    & 0.693    & 0.723   \\
\textbf{MSA-AUD}   & \textbf{0.694}   & \textbf{0.892}   & \textbf{0.600}   & \textbf{0.661}    & \textbf{0.721}   & \textbf{0.713}   & \textbf{0.962}  & \textbf{0.969}  & \textbf{0.919}  & \textbf{0.632}    & \textbf{0.714}    & \textbf{0.775}   \\
\noalign{\smallskip}\hline\noalign{\smallskip}
\end{tabular}
\label{main}
\end{table*}

\subsubsection{Baseline}
We compare our method with MSA-Real, JbDA(\textbf{J}acobian-\textbf{b}ased \textbf{D}ataset \textbf{A}ugmentation), T-RND(\textbf{T}argeted \textbf{ran}domly), and Random for benchmarking purposes.
\begin{itemize}[leftmargin=*]
\item \textbf{MSA-Real}~\cite{shen2022model, wu2022model} utilized only real data to train the clone model. The previous works implemented model stealing attacks against node classification in this way. It serves as a benchmark without data generation and is used to assess whether synthetic samples improve attack performance.

\item \textbf{JbDA}~\cite{papernot2017practical} employs a data augmentation approach based on the Jacobian matrix to generate images. Specifically, JbDA calculates the gradient on the original image and adds a small noise in the direction of the gradient. The modified image is then fed into the target model as queries. 

\item \textbf{T-Rnd}~\cite{juuti2019prada} is an improvement over JbDA, which switches from single-step perturbation to multi-step perturbation.
In each perturbation step, T-Rnd algorithm chooses a target class at random and then computes the Jacobian matrix of the output that the $\mathcal{M}_{C}$ produces in this class. 
Furthermore, T-Rnd introduces smaller noise along the gradient direction compared to JbDA when adding it to the original sample.

JbDA and T-RND were originally designed for continuous inputs (images). To adapt them for the graph domain, we first relax the adjacency matrix into continuous variables, add a small noise along the gradient direction, and finally, restore the adjacency matrix to discrete variables.

\item \textbf{Random}~\cite{roberts2019model} utilizes random images to steal the target model.
We migrate it to the graph data and randomly generate some graphs satisfying the real graph distribution to expand the initial dataset.
The number of random samples in MSA-Rnd is the same as that in other attack methods based on synthesizing samples.

\end{itemize}

\subsubsection{Implementation details}
In our study, we construct the target model as a Graph Neural Network (GNN) comprising three hidden layers, with each GNN layer containing 128 units. We adopt the average function as the pooling mechanism for the GNN layers. 
We employ the ReLU activation function and Adam optimizer with a learning rate of 0.01.
We use the cross-entropy loss to train the target model and the clone model.
The number of rounds of generating samples (N) is set to 20. The proportions of the modified adjacency matrix ($\alpha$) and mixed nodes ($\gamma$) are set to 0.05 and 0.1, respectively.
Lastly, we report the average fidelity of 5 runs for each model-stealing attack method.

\subsection{Attack Performance on Various Datasets}
\label{mainre}
In this section, we conduct experiments on four datasets: ENZYMES, COIL-DEL, NCI1, and TRIANGLES. The upper part of Table \ref{main} represents attack methods that do not modify node features, while the lower part includes attack methods that modify both node features and adjacency matrix. From the experimental results presented in Table \ref{main}, we draw the following conclusions:

\textit{Finding 1}: The generated samples help to alleviate the issue of insufficient real data. Based on the results in the table, all methods that rely on generating new samples have shown a significant performance boost compared to MSA-Real. For instance, on the COIL-DEL dataset under the SAGE model, JbDA, T-RND, MSA-AU, Random, MSA-AD, and MSA-AUD have demonstrated improvements of 24.9\%, 23.7\%, 36.2\%, 26.9\%, 42.2\%, and 43.3\% respectively over MSA-Real. This suggests that the generated samples can, to some extent, replace real samples to guide clone model updates.

\textit{Finding 2}: Considering the uncertainty in the process of synthesizing samples can effectively improve the attack performance of the method. Compared to all other methods that solely modify the adjacency matrix, including MSA-Real, JbDA, and T-RND, MSA-AU outperforms other attack methods across all datasets and target models. Notably, on the TRIANGLES dataset under the GIN model, MSA-AU even surpasses the performance of MSA-AD and Random, two methods that modify both node features and adjacency matrix. This indicates that the samples generated by MSA-AU indeed have high values to be queried. MSA-AUD, which combines the advantages of MSA-AD and MSA-AU, simultaneously considers authenticity, uncertainty, and diversity in the process of synthesizing new samples, and shows the best performance.

\textit{Finding 3}: 
The difficulty of model stealing attacks is strongly correlated with the number of classes in the dataset. The improvements over MSA-Real achieved by other attack methods exhibit significant variations across different datasets. On the NCI1 dataset, the improvements over MSA-Real are relatively modest, even with our method MSA-AUD achieving only a 10\% relative improvement. This is primarily attributed to the NCI1 dataset's limited number of classes (2 classes), making it more susceptible to model stealing attacks. Even with a small number of real samples, attackers can achieve sufficiently good stealing performance.
Conversely, on the COIL-DEL dataset with 100 classes, MSA-Real only achieves around 20\% stealing performance, while our methods can enhance the stealing performance to around 70\%. Consequently, we will conduct further analytical experiments on the COIL-DEL dataset to observe the performance of different attack algorithms in various scenarios.

\begin{table*}[]
\centering
\caption{Imperceptibility of different attack methods. The best results are highlighted in bold.} 
\begin{tabular}{l|cccccc}
\hline\noalign{\smallskip}
Algorithm & Degree Distribution & \#Triangles    & Clustering   Coefficient & Transitivity & \#Cliques  & Node Connectivity \\
\noalign{\smallskip}\hline\noalign{\smallskip}
JbDA      & 0.11652874          & 11.55641026 & 0.076869                 & 0.066009     & 4.807692 & 0.551384          \\
T-RND      & 0.22114451          & 160.6102564 & 0.050915                 & 0.056535     & 44.51795 & 2.867363          \\
\textbf{MSA-AU}    & \textbf{0.02605779}          & \textbf{4.074358970}  & \textbf{0.019472}                 & \textbf{0.015664}     & \textbf{1.676923} & \textbf{0.209449}          \\
\noalign{\smallskip}\hline\noalign{\smallskip}
Random    & 0.13860179          & 99.33076923 & 0.133064                 & 0.139343     & 32.78974 & 3.930681          \\
\textbf{MSA-AD}     & \textbf{0.00002089}          & \textbf{0.017948720}  & \textbf{5.33E-05}                 & \textbf{5.63E-05}     & \textbf{0.015385} & \textbf{0.000809}          \\
MSA-AUD     & 0.02348852          & 2.917948720  & 0.017514                 & 0.012399     & 1.471795 & 0.169263         \\
\noalign{\smallskip}\hline\noalign{\smallskip}
\end{tabular}
\label{real}
\end{table*}

\begin{figure*}[h]
  \centering
  \includegraphics[width=\linewidth]{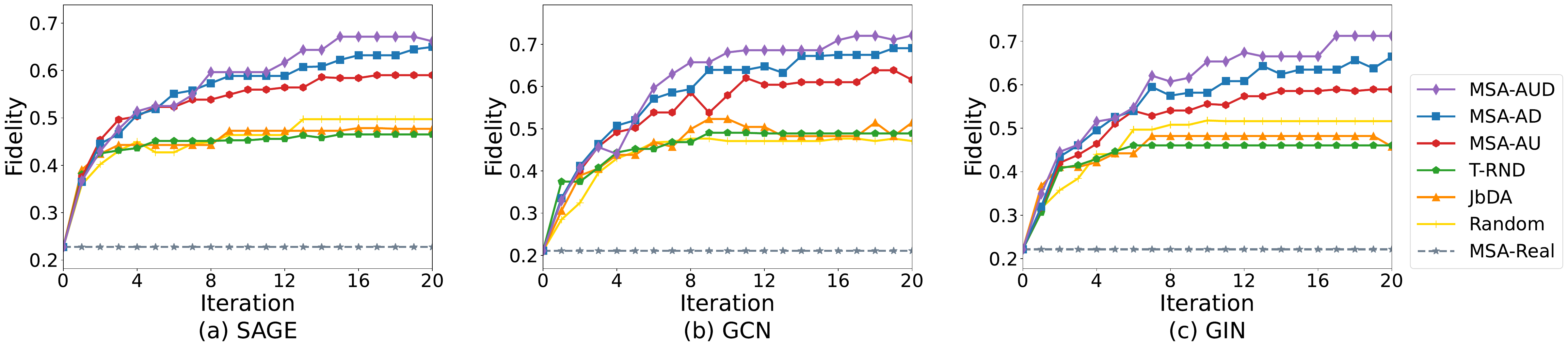}
  \caption{Attack performance with the different number of query samples.}
  \label{epoch}
\end{figure*}

\subsection{Imperceptibility of Different Attack Methods}
\label{imper}
In this section, we conduct an analysis of the imperceptibility of attack methods. Table \ref{real} presents the discrepancies in statistical measures between graphs generated using various attack methods and the statistical measures of the original graphs. These statistical measures include degree distribution, triangle counts, clustering coefficient, transitivity, number of cliques, and average node connectivity. Smaller discrepancies indicate a higher level of authenticity in the generated graphs. From Table \ref{real}, it can be observed that among all methods that only modify the adjacency matrix, MSA-AU exhibits the best imperceptibility due to its adherence to authenticity constraints. In addition, MSA-AD, as it only mixes a small amount of the topological structure from other real graphs, yields the highest level of authenticity in all methods. MSA-AUD's authenticity falls between MSA-AD and MSA-AU, demonstrating significantly improved authenticity compared to the baselines. Random, characterized by a highly random graph generation process, exhibits the lowest level of imperceptibility. 

It is pertinent to note that we did not assess MSA-AD's imperceptibility at the node feature level in this study. This is due to MSA-AD's feature distribution, which should be consistent with the original dataset since it just transfers node features and does not create new ones.

\subsection{Performance with Different Knowledge}
\label{knowledge}

\begin{table*}[]
\centering
\caption{Attack performance with different model architectures. The best results are highlighted in bold.} 
\begin{tabular}{l|ccc|ccc|ccc}
\hline\noalign{\smallskip}
Victim Model & \multicolumn{3}{c|}{SAGE} & \multicolumn{3}{c|}{GCN} & \multicolumn{3}{c}{GIN} \\
\noalign{\smallskip}\cline{1-10} \noalign{\smallskip}
Clone Model   & SAGE    & GCN     & GIN     & SAGE     & GCN     & GIN     & SAGE   & GCN    & GIN     \\
\noalign{\smallskip}\cline{1-10} \noalign{\smallskip}
MSA-Real   & 0.2276 & 0.1109 & 0.1333 & 0.2939 & 0.2109 & 0.1638 & 0.2458 & 0.1439 & 0.2215 \\
JbDA   & 0.4769 & 0.2673 & 0.3160 & 0.4282 & 0.5147 & 0.4288 & 0.4016 & 0.3343 & 0.4577 \\
T-RND   & 0.4651 & 0.2590 & 0.3099 & 0.4298 & 0.4888 & 0.4321 & 0.4061 & 0.3487 & 0.4606 \\
\textbf{MSA-AU} & \textbf{0.5901} & \textbf{0.3083} & \textbf{0.3362} & \textbf{0.5032} & \textbf{0.6157} & \textbf{0.5250} & \textbf{0.4087} & \textbf{0.3904} & \textbf{0.5894} \\
\noalign{\smallskip}\cline{1-10} \noalign{\smallskip}
Random & 0.4971 & 0.1109 & 0.1833 & 0.5013 & 0.4708 & 0.4692 & 0.3288 & 0.2590 & 0.5163 \\
MSA-AD  & 0.6497 & 0.2674 & 0.3106 & 0.5984 & 0.6907 & 0.6173 & 0.4497 & 0.4317 & 0.6647 \\
\textbf{MSA-AUD}  & \textbf{0.6615} & \textbf{0.2869} & \textbf{0.3340} & \textbf{0.6449} & \textbf{0.7212} & \textbf{0.6631} & \textbf{0.4740} & \textbf{0.4609} & \textbf{0.7125} \\
\noalign{\smallskip}\hline\noalign{\smallskip}
\end{tabular}
\label{unware}
\end{table*}

\begin{figure*}[h]
  \centering
  \includegraphics[width=\linewidth]{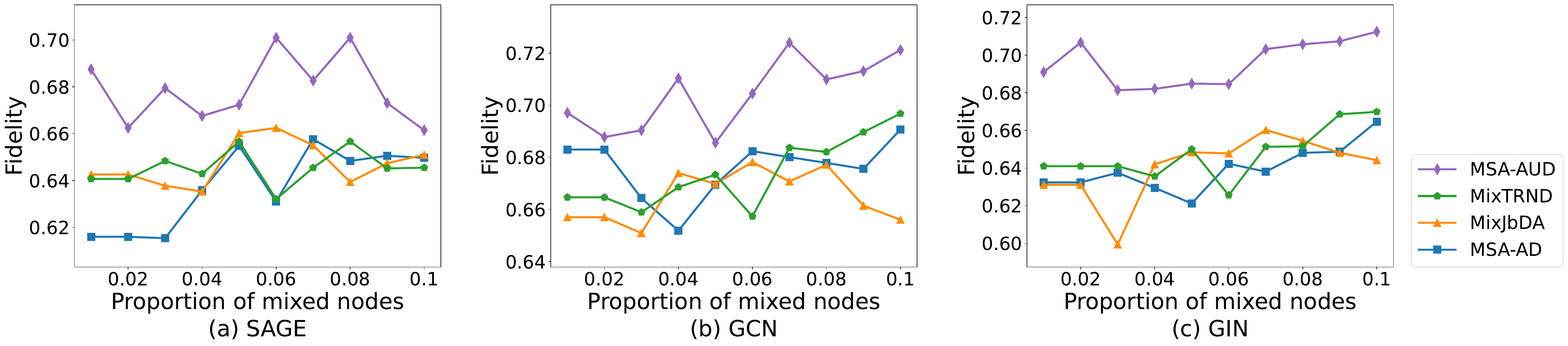}
  \caption{Attack performance with different proportions of mixed nodes.}
  \label{augsize}
\end{figure*}

\subsubsection{Query budget}
\label{querybud}
We present the performance variations of different attack methods during the data generation process in Fig \ref{epoch}. The number of queries we have is equivalent to the number of samples in the pool. The observations are as follows:

\textit{Finding 1}:
With an increase in the query count, there is a noticeable improvement in the performance of all methods. Particularly, in the initial rounds, the performance gains for the attack methods are particularly prominent. On the SAGE model, all attack methods achieve a fidelity improvement exceeding 20\% within the first four rounds. The rate of fidelity improvement gradually diminishes as the rounds progress, eventually leading to fidelity convergence.

\textit{Finding 2}:
Across all models, our method consistently demonstrates optimal performance. In the initial rounds, the performance of MSA-AUD and MSA-AU is closely matched, but with the advancement of iteration rounds, MSA-AUD progressively exhibits significantly superior performance compared to MSA-AU. This observation underscores the effectiveness of integrating MSA-AD in alleviating the problem of lack of diversity in the generated data caused by the convergence of the clone model in MSA-AU.

\subsubsection{Unware model architecture}
\label{unware_exp}
In this part, we assume that the attacker lacks knowledge of the target model's architecture. The experimental results presented in Table \ref{unware} yield the following insights:

\textit{Finding 1}:
MSA-AUD and MSA-AU demonstrate superior performance compared to other methods, validating the effectiveness of our approach. Even when the attacker is unaware of the target model's structure, our method can still achieve a certain level of attack performance. 

\textit{Finding 2}:
When the category of the target model is unknown, the performance of all attack methods experiences a decline. This experimental result aligns with intuition, suggesting that safeguarding the target model against stealing attacks can begin by maintaining secrecy regarding the model's category.

\subsection{Parametric Analysis}
\label{param}
In this section, we further analyze the effect of the proportion of mixed nodes, proportion of modifications on attack performance.

\subsubsection{Proportion of mixed nodes($\gamma$)}
\label{mix}
Fig \ref{augsize} illustrates the performance variations of different attack methods when increasing the proportion of mixed nodes. To rigorously assess the efficacy of MSA-AU, we integrate JbDA and T-Rnd with MSA-AD, resulting in MixJbDA and MixTRND, respectively. It is discernible that:

\textit{Finding 1}:
MSA-AUD consistently exhibits superior performance across all scenarios. While MixJbDA and MixTRND show marginal improvements in some instances when combined with MSA-AD, these enhancements remain relatively inconspicuous. Furthermore, the integration of MSA-AD with JbDA and T-RND, given their inherently low authenticity, paradoxically tends to compromise the imperceptibility of MSA-AD.

\textit{Finding 2}:
Elevating the mixing ratio generally leads to improved MSA-AD performance in most cases. However, MSA-AUD, MixJbDA, and MixTRND, due to their incorporation of other methods, are less influenced by changes in the mixing ratio.

\begin{figure*}[h]
  \centering
  \includegraphics[width=\linewidth]{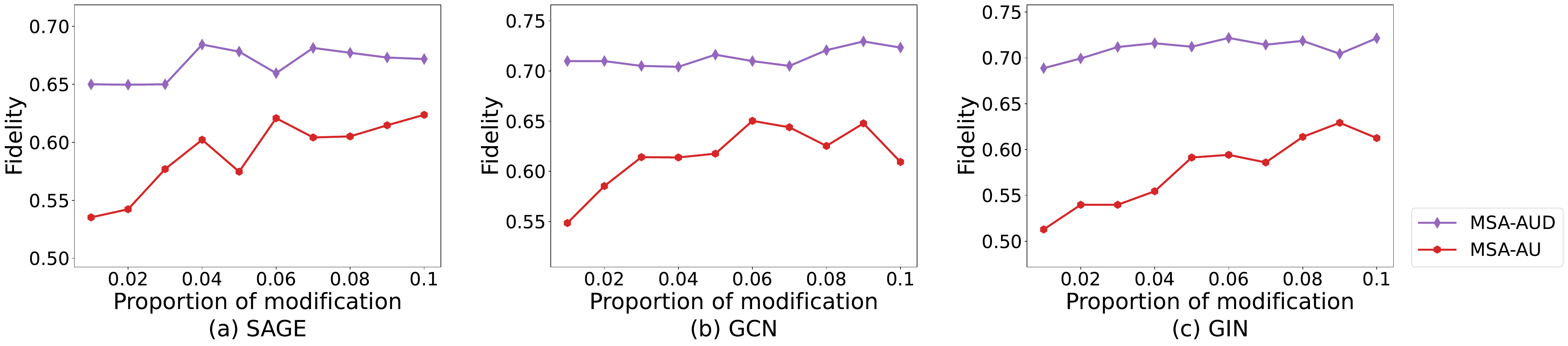}
  \caption{Attack performance with different proportions of modifications on adjacency matrix.}
  \label{cos}
\end{figure*}

\begin{figure*}[h]
  \centering
  \includegraphics[width=\linewidth]{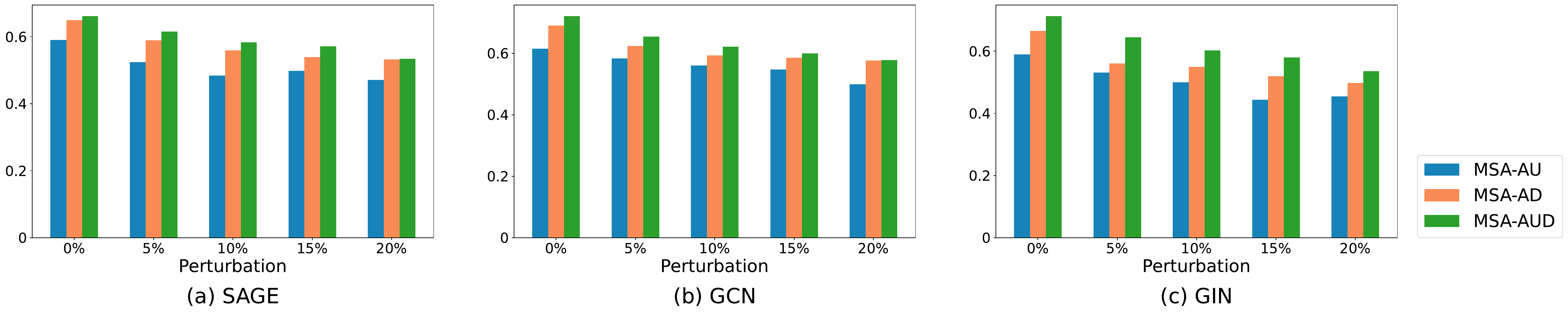}
  \caption{Attack performance with perturbations of different scales added to the target model's output.}
  \label{def}
\end{figure*}

\begin{table}[htbp]
\centering
\caption{The detection accuracy of the defender.}
\begin{tabular}{l|ccc}
\noalign{\smallskip}\hline\noalign{\smallskip}
Attack algorithm & SAGE   & GCN    & GIN    \\
\noalign{\smallskip}\hline\noalign{\smallskip}
MSA-AU    & 0.5176 & 0.5208 & 0.5401 \\
MSA-AD     & 0.5337 & 0.5288 & 0.5593 \\
MSA-AUD     & 0.5176 & 0.5032 & 0.5288 \\
\noalign{\smallskip}\hline\noalign{\smallskip}
\end{tabular}
\label{accc}
\end{table}

\subsubsection{Proportion of modifications on adjacency matrix($\alpha$)}
\label{Modified_Edge_Numbers}
In this part, we investigate the influence of the proportions of modifications on adjacency matrix on MSA-AU and MSA-AUD. The experimental results presented in Fig \ref{cos} reveal several noteworthy findings.

\textit{Finding 1}:
Firstly, both MSA-AU and MSA-AUD exhibit improved performance when a greater number of modifications are applied. Notably, the performance of MSA-AU appears to be nearly proportional to the number of modified edges. Conversely, MSA-AUD's performance is less evidently affected by the number of modifications. This phenomenon arises from MSA-AUD's incorporation of MSA-AD, which introduces a dual influence on its performance. Therefore, MSA-AUD is not significantly affected by the parameters of a certain strategy, aligning with the findings from the experiments in Section \ref{mix}.

\textit{Finding 2}:
It is important to highlight that even when modifying only 1\% of the adjacency matrix, both MSA-AU and MSA-AUD exhibit noteworthy model stealing capabilities. This underscores the resilience and effectiveness of our approach, demonstrating its proficiency in model stealing attacks, even in situations involving a relatively low percentage of modifications.

\subsection{Defense}
\label{defense}
In this section, we analyze how our method performs under existing defense strategies. We tested two specific strategies to counter our attacks. One strategy involves adding noise to the model's output~\cite{shi2017evasion, orekondy2019prediction}, while the other relies on the detection of generated graphs~\cite{ma2023generated}.

Specifically, we introduce random noise into the responses from the target model, which induces the clone model to learn predictions that are inconsistent with those of the target model. Note that we operate under the assumption that attackers can only access the hard label returned by the target model. Consequently, defenders are compelled to modify the predictions of the target model if they wish to introduce noise. As the level of noise increases, the confidence in the predictions made by the target model diminishes. The attack performance of our three methods under varying perturbations is presented in Fig \ref{def}. Our findings reveal that the noise-based defense strategy does not mitigate our attack threat. Even with the addition of 20\% noise, MSA-AUD's model stealing performance remains above 50\% for all three models.

Furthermore, we employ state-of-the-art detection mechanisms to distinguish between our generated graphs and real graphs, which is reported in Table \ref{accc}. Notably, even these advanced detection mechanisms struggle to discern the differences between our generated graphs and real ones. This is attributed to two primary reasons. First, our generation process considers imperceptibility, resulting in minimal differences between the generated and real graphs. Second, existing detection methods are designed to differentiate between real graphs and generated graphs(e.g. by generators like GANs or diffusion models).However, our generated graphs are minimally changed from real ones, and there is currently no reliable way to effectively distinguish them from genuine graphs.

\subsection{Discussion}
\label{discuss}
Based on our experimental findings (Section \ref{defense}), existing defense strategies are not effective against our attack methods. To address this, we propose three potential defense directions:
\begin{itemize}
    \item Section \ref{unware_exp} suggests that maintaining the secrecy of the model's architecture and training details is a straightforward and effective defense strategy, as the attack performance degrades when the model's architecture is unknown.
    \item As shown in Section \ref{querybud}, the attacker's success depends on their query budget. Increasing the cost of these queries is a viable defense option.
    \item In Section \ref{param}, we observed that attackers perform better when they introduce larger perturbations to the data. While current detection mechanisms cannot differentiate between synthetic and genuine graphs, future research could focus on building more robust detectors based on the attacker's sample generation (e.g. adversarial training).
\end{itemize}
\textbf{Ethics statement.} The datasets utilized in our study are all open-source, and the target model is locally trained, thereby eliminating any risk of privacy breaches. Our research on model stealing attacks in the context of graph classification is motivated by the desire to encourage the community to design more secure graph classification models.
\section{Related Work}
\textbf{Graph Neural Networks}
Graphs are ubiquitous in our lives~\cite{cook2006mining, sakr2011graph, vishveshwara2002protein}.
For instance, subway tracks in a city form a transportation graph~\cite{derrible2009network}.
The social network is comprised of the relationship between online users~\cite{viegas2004social}.
Due to the expressiveness of graphs, graph analysis problems have attracted increasing attention in recent years.
When compared to DeepWalk~\cite{perozzi2014deepwalk}, node2vec~\cite{grover2016node2vec} and other traditional node embedding methods~\cite{qiu2018network, tang2015line, tang2015pte, cui2018survey}, GNNs~\cite{kipf2016semi, hu2020gpt, hamilton2017inductive} are far more effective, hence they have found widespread application in graph analysis jobs.

GNNs can typically be divided into two families: spectral methods~\cite{li2018adaptive, henaff2015deep} and spatial methods~\cite{zhu2018modelling, gilmer2017neural}. 
Bruna et al.~\cite{bruna2013spectral} firstly generalize the convolution operation from Euclidean data to non-Euclidean data.
GCN~\cite{kipf2016semi}, one of the most popular algorithms in this family, simplifies the GNN model using ChebNet\cite{defferrard2016convolutional}'s first-order approximation.
There are numerous GCN-based variants~\cite{du2017topology, chen2018fastgcn, tian2020ra}.
For instance, TAGCN~\cite{du2017topology} uses multiple kernels to extract neighborhood information of different receptive fields, whereas GCN uses a single convolution kernel.
RGCN~\cite{tian2020ra} extended GCN to accommodate heterogeneous graphs with different types of nodes and relations.
GraphSage~\cite{hamilton2017inductive} is a famous spatial approach that utilizes various aggregation functions to combine node neighborhood information.
It is inductive and can predict freshly inserted nodes.
In addition, GraphSage restricted the amount of aggregated neighbors using sampling to be able to process large graphs.
GAT~\cite{tian2020ra} included the attention mechanism in its architecture. 
In each graph convolutional operation, distinct attention scores are learned for each of the target node's neighbors.

In recent years, GNNS have been pointed out to be threatened by adversarial attacks~\cite{zhang2020gnnguard, ma2020towards}, attribute inference attacks~\cite{wu2021adapting}, and model stealing attacks~\cite{shen2022model, wu2022model}. More details can be found in the recent surveys~\cite{wu2020comprehensive, zhou2020graph}.

\noindent \textbf{Model Stealing Attack}
An increasing number of privacy attacks~\cite{rigaki2020survey, song2020information, wu2016methodology, fredrikson2015model} have been proposed to violate the owner's intellectual property from a data and model perspective.
Unlike adversarial attacks~\cite{finlayson2019adversarial, madry2017towards, dong2018boosting}, which try to undermine the performance and credibility of the target model, privacy attacks aim to violate the target model's privacy by abusing its permissions.
Model stealing attack~\cite{tramer2016stealing, miura2021megex, yoshida2019model, takemura2020model}, which steals various components of a black-box machine learning(ML) model(e.g. hyperparameters~\cite{wang2018stealing}, architecture~\cite{oh2019towards}), is one of the most common privacy attacks.
Most recent works are mostly concerned with stealing the functionality~\cite{jagielski2020high, orekondy2019knockoff} of the target model.
To be specific, the attacker anticipates constructing a good clone of the target model($\mathcal{M}_{T}$) through $\mathcal{M}_{T}$'s query API.
The clone model can be further used in other privacy attacks(e.g. membership inference attack~\cite{shokri2017membership, hu2022membership, choquette2021label}).

With query access, Tramer et al.~\cite{tramer2016stealing} proposed a model stealing attack against a machine learning model.
Since then, researchers in fields like computer vision~\cite{wang2022black, yu2020cloudleak}, generative adversarial networks~\cite{hu2021stealing}, and recommendation systems~\cite{yue2021black} have explored model stealing attacks.
Traditional model stealing attacks assume that the attacker can obtain sufficient data to achieve their goal.
Papernot et al.~\cite{papernot2017practical} established a framework with limited access to the data distribution sample set.
They perturbed the original samples to progressively obtain synthetic examples for training the clone model.
The defense of model stealing attack is mainly divided into two categories. One is to detect whether the target model is under stealing attack~\cite{jia2021entangled, szyller2021dawn}, and the other is to prevent in advance by adding noise and other ways to destroy the availability of the clone model~\cite{shi2018active, alabdulmohsin2014adding}.

There have been few works~\cite{shen2022model, wu2022model} that focus on stealing the graph neural networks.
Yun et al.~\cite{shen2022model} proposed different attacks based on the attacker’s background knowledge and the responses of the target models.
They assumed that the target model could provide node embedding, prediction, or t-sne projection vectors.
Bang Wu.~\cite{wu2022model} assumed that it was possible to obtain a portion of the training graph and a shadow graph with the same distribution as the training graph.
All of these works, however, focus on node classification while ignoring the graph-level task.
\section{Conclusion and Future Work}
In conclusion, recent research has highlighted GNNs' vulnerability to model-stealing attacks, mainly focused on node classification tasks, and questioned their practicality. We propose strict settings using limited real data and hard-label awareness for synthetic data generation, simplifying the theft of the target model. We introduce three model-stealing methods for various scenarios: MSA-AU emphasizes uncertainty, MSA-AD adds diversity, and MSA-AUD combines both.
Our experiments consistently show the effectiveness of our methods in enhancing query efficiency, and improving model-stealing performance. In the future, we will design more effective defense mechanisms to address the threats posed by model stealing attacks.

\clearpage

\bibliographystyle{plain}
\bibliography{msagnn}
\clearpage
\appendix

\section{Appendix}

\begin{table*}[htbp]
\centering
\caption{Attack performance with different pooling functions. The best results are highlighted in bold.} 
\begin{tabular}{l|ccc|ccc|ccc}
\hline\noalign{\smallskip}
Pooling Function & \multicolumn{3}{c|}{Average} & \multicolumn{3}{c|}{Sum} & \multicolumn{3}{c}{Maximum} \\
\noalign{\smallskip}\cline{1-10} \noalign{\smallskip}
Model   & SAGE    & GCN     & GIN     & SAGE     & GCN     & GIN     & SAGE   & GCN    & GIN     \\
\noalign{\smallskip}\cline{1-10} \noalign{\smallskip}
MSA-Real   & 0.2276 & 0.2109 & 0.2215 & 0.1923 & 0.1083 & 0.0667 & 0.3196 & 0.0785 & 0.1644 \\
JbDA   & 0.4769 & 0.5147 & 0.4577 & 0.4391 & 0.5051 & 0.4199 & 0.4792 & 0.4423 & 0.4096 \\
T-RND   & 0.4651 & 0.4888 & 0.4606 & 0.4285 & 0.5330  & 0.4131 & 0.4625 & 0.4337 & 0.3782 \\
\textbf{MSA-AU} & \textbf{0.5901} & \textbf{0.6157} & \textbf{0.5894} & \textbf{0.5224} & \textbf{0.6378} & \textbf{0.5580}  & \textbf{0.5776} & \textbf{0.5833} & \textbf{0.4766} \\
\noalign{\smallskip}\cline{1-10} \noalign{\smallskip}
Random & 0.4971 & 0.4708 & 0.5163 & 0.4407 & 0.5298 & 0.4740  & 0.4958 & 0.4737 & 0.4407 \\
MSA-AD  & 0.6497 & 0.6907 & 0.6647 & 0.5968 & 0.6606 & 0.5731 & 0.6426 & 0.6385 & 0.5250  \\
\textbf{MSA-AUD}  & \textbf{0.6615} & \textbf{0.7212} & \textbf{0.7125} & \textbf{0.5827} & \textbf{0.6978} & \textbf{0.6551} & \textbf{0.6792} & \textbf{0.6801} & \textbf{0.5869} \\
\noalign{\smallskip}\hline\noalign{\smallskip}
\end{tabular}
\label{pool}
\end{table*}

\begin{table*}[]
\centering
\caption{Attack performance with different uncertainty metrics. The best results are highlighted in bold.} 
\begin{tabular}{l|ccc|ccc|ccc}
\hline\noalign{\smallskip}
Target Model & \multicolumn{3}{c|}{SAGE} & \multicolumn{3}{c|}{GCN} & \multicolumn{3}{c}{GIN} \\
\noalign{\smallskip}\cline{1-10} \noalign{\smallskip}
Metric & Margin    & Max     & Entropy     & Margin    & Max     & Entropy  & Margin    & Max     & Entropy     \\
\noalign{\smallskip}\cline{1-10} \noalign{\smallskip}
MSA-AU & 0.5901 & 0.6019 & 0.6019 & 0.6157 & 0.6083 & 0.5958 & 0.5894 & 0.5808 & 0.6000    \\
\textbf{MSA-AUD}  & \textbf{0.6615} & \textbf{0.6715} & \textbf{0.6798} & \textbf{0.7212} & \textbf{0.7186} & \textbf{0.7016} & \textbf{0.7125} & \textbf{0.7269} & \textbf{0.6933}    \\
\noalign{\smallskip}\hline\noalign{\smallskip}
\end{tabular}
\label{unmetric}
\end{table*}

\subsection{Parameter Analysis}
\subsubsection{Pooling function}
In this section, we employ various pooling functions, namely average pooling, maximum pooling, and sum pooling, to systematically evaluate their impact on the overall performance of our algorithm. From the experimental results Table \ref{pool}, it can be observed that our algorithm consistently outperforms other methods across all the pooling functions considered. 

It is notable that the choice of pooling function has a more pronounced impact on the performance of MSA-Real compared to its effect on our attack techniques. As an illustrative example, when targeting the GIN attack model, MSA-Real achieves a performance boost of 15.5\% when the pooling function is set to 'Average', compared to 'Sum'. However, in the case of MSA-AU and MSA-AUD, the performance improvements are only 3.14\% and 5.74\%, respectively. These results underscore the robustness of the our algorithms, indicating that it remains unaffected by the choice of pooling function.

\subsubsection{Uncertainty metric}
In this section, we delve into the impact of employing different uncertainty metrics, namely margin of confidence (Margin), maximum confidence (Max) and entropy of confidence vecter (Entropy), on MSA-AU and MSA-AUD. The experimental results presented in Table \ref{unmetric} shed light on several crucial observations. On one hand, it is evident that the choice of uncertainty metrics has a relatively minimal impact on the performance of our approach. Regardless of what the metric is employed, our method consistently demonstrates robust attack capabilities. On the other hand, under varying uncertainty metrics, MSA-AUD consistently outperforms MSA-AU, which is attributed to the integration of diversity in the data generation of MSA-AUD.

\end{document}